\RequirePackage{amsmath}
\documentclass[runningheads]{llncs}
\usepackage{hyperref}
\hypersetup{
    colorlinks=true,
    linkcolor=blue,
    filecolor=magenta,      
    urlcolor=cyan,
}
\usepackage{graphicx}
\usepackage{amsmath,amssymb} % define this before the line numbering.
\usepackage{color}
\usepackage[caption=false]{subfig}
\begin{document}
\title{A Semi-Supervised Data Augmentation Approach using 3D Graphical Engines} 
% Replace with your title

\titlerunning{A Semi-Supervised Data Augmentation Approach}
% Replace with a meaningful short version of your title
%
\author{Shuangjun Liu \and Sarah Ostadabbas}
%
%Please write out author names in full in the paper, i.e. full given and family names. 
%If any authors have names that can be parsed into FirstName LastName in multiple ways, please include the correct parsing, in a comment to the volume editors:
%\index{Lastnames, Firstnames}
%(Do not uncomment it, because you may introduce extra index items if you do that, we will use scripts for introducing index entries...)
\authorrunning{S. Liu and S. Ostadabbas}
% Replace with shorter version of the author list. If there are more authors than fits a line, please use A. Author et al.
%

\institute{Augmented Cognition Lab, Electrical and Computer Engineering Department,\\
	Northeastern University, Boston, USA\\
\email{\{shuliu,ostadabbas\}@ece.neu.edu}\\
\url{http://www.northeastern.edu/ostadabbas/}}

\newcommand{\figPhysicsBased}{
\begin{figure}[t]
    \centering
     \vspace{-.1in}
    \includegraphics[width=0.99\textwidth]{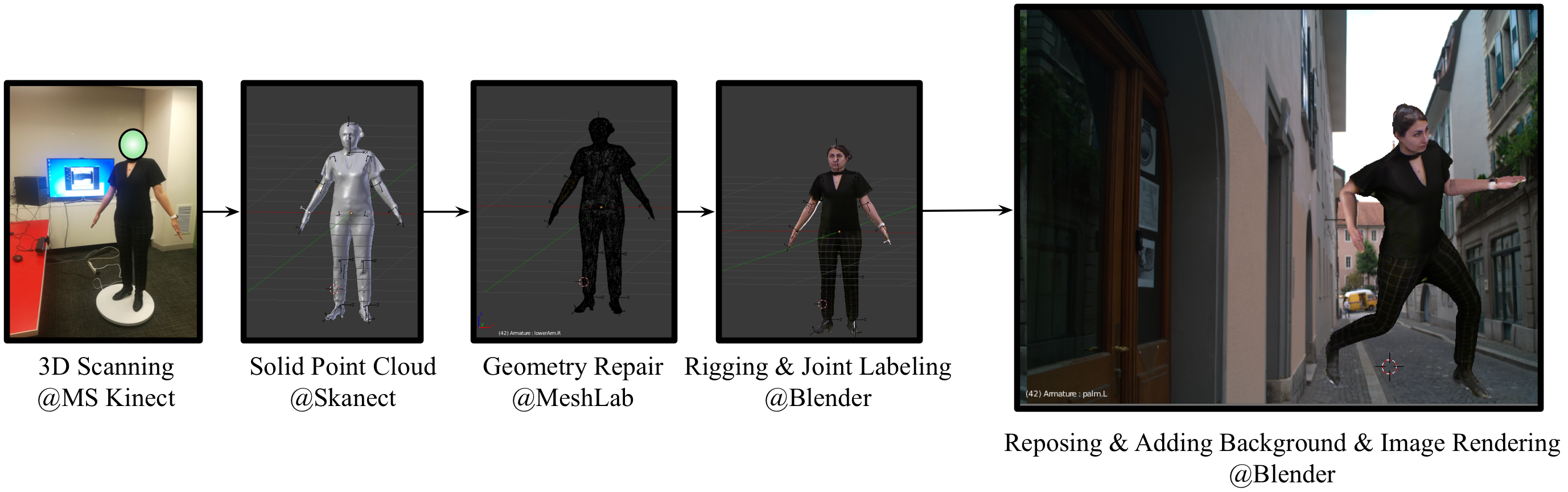}
    \caption{\fontsize{9}{11}\selectfont An overview of our pipeline on the physically-valid semi-supervised human pose data augmentation approach that leads to forming ScanAva datasets.}
    \vspace{-.1in}
    \label{fig:physicsBased}
    \vspace{-.1in}
\end{figure}
}

\newcommand{\figScan}{
\begin{figure}[t]
 \centering
  \vspace{-.1in}
  \subfloat[]{\label{fig:skeleton}\includegraphics[width=0.25\linewidth,{trim=0in 0in 0in 0in,
  clip=true}]{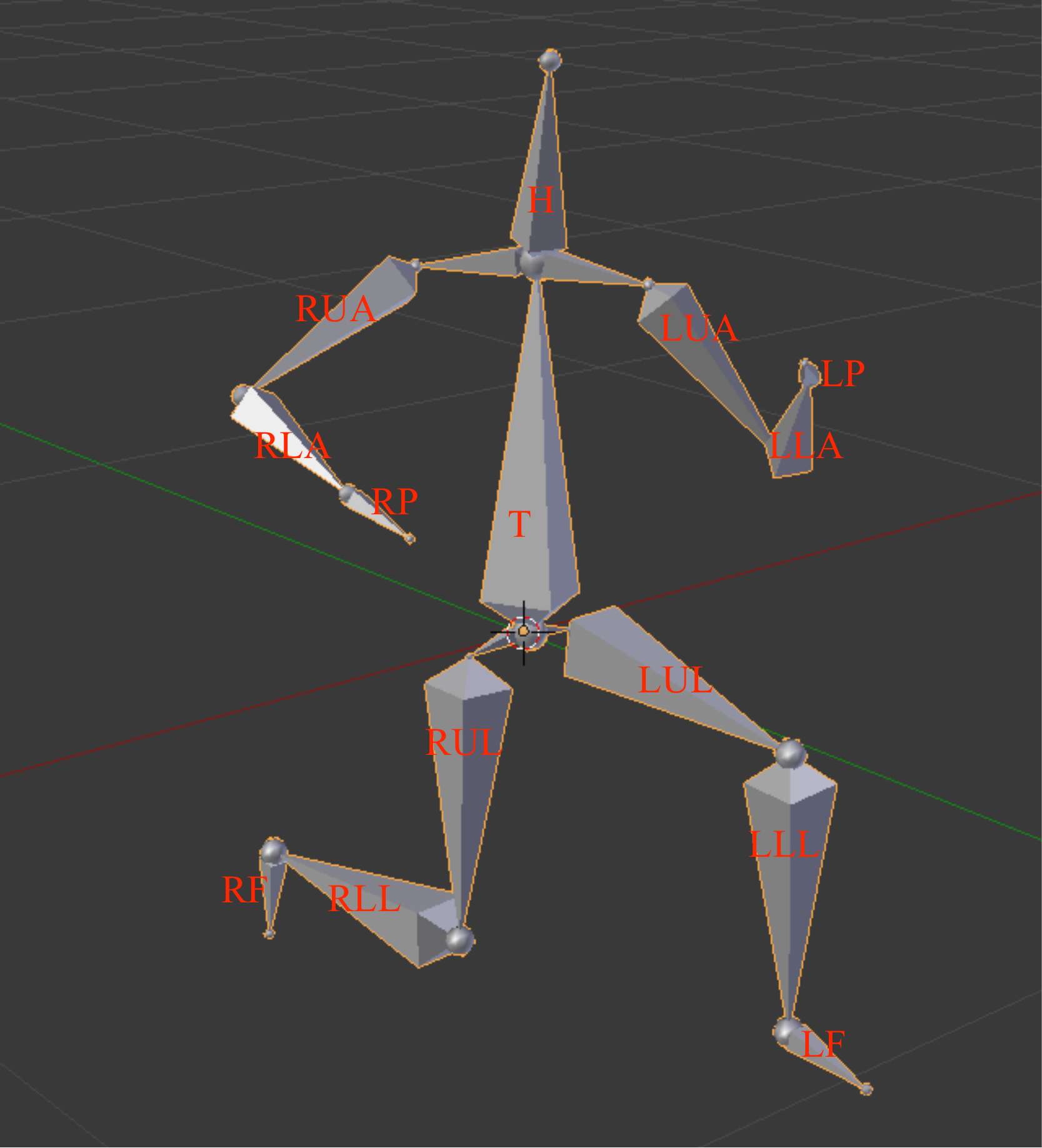}}
 \subfloat[]{\label{fig:scanProcess}\includegraphics[width=0.4\linewidth, trim=0in 0in 0in 0in,
  clip=true]{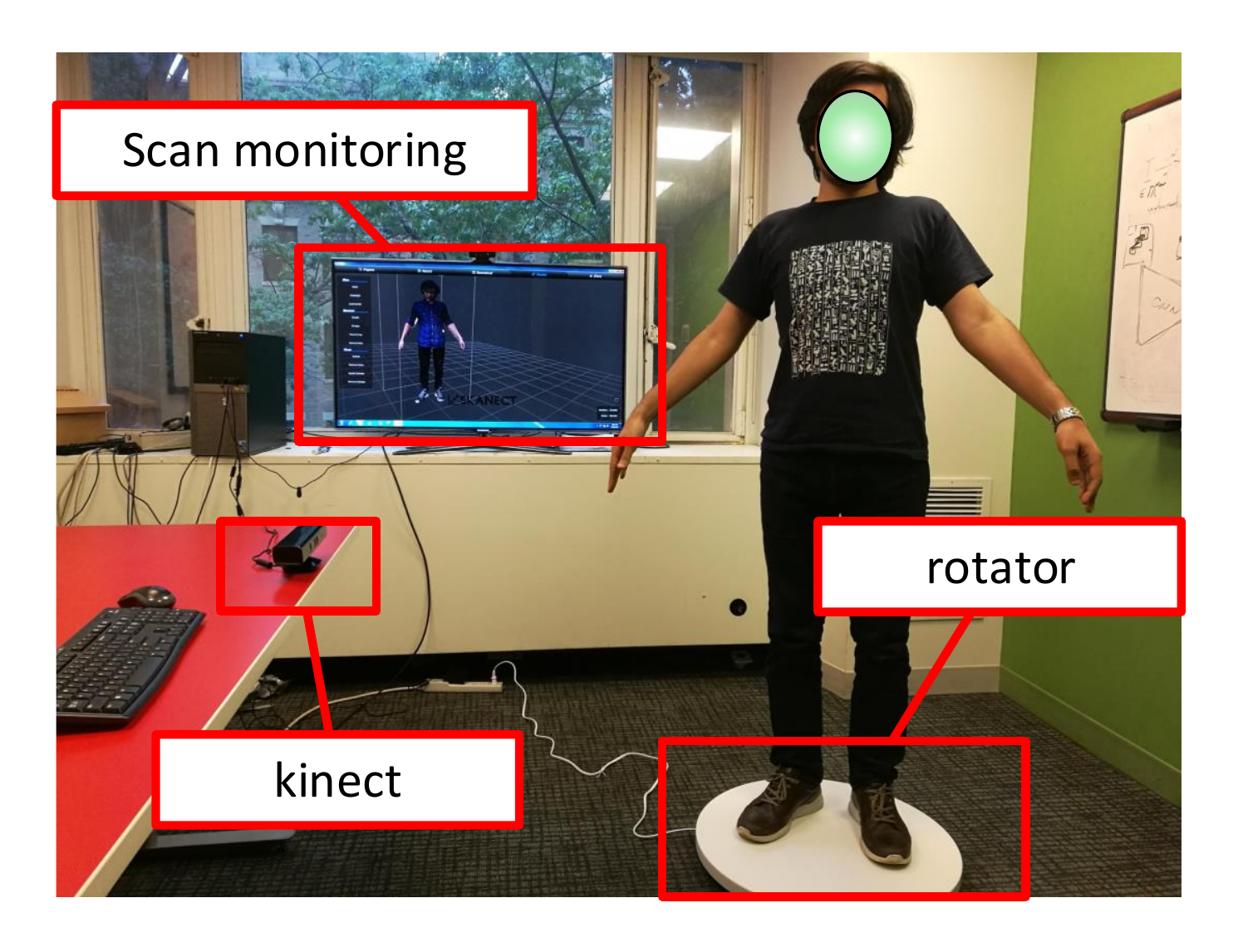}}
  \caption{\fontsize{9}{11}\selectfont  (a) The human skeleton with 14 limbs and corresponding joints, (b) Human body 3D scanning procedure.}
        \vspace{-.1in}
\label{fig:scan}
    \vspace{-.1in}
\end{figure}
}

\newcommand{\figRigged}{
\begin{figure}[t]
 \centering
  \vspace{-.1in}
 \subfloat[]{\label{fig:riggedModel}\includegraphics[width=0.435\linewidth, trim=0in 0in 0in 0in,
  clip=true]{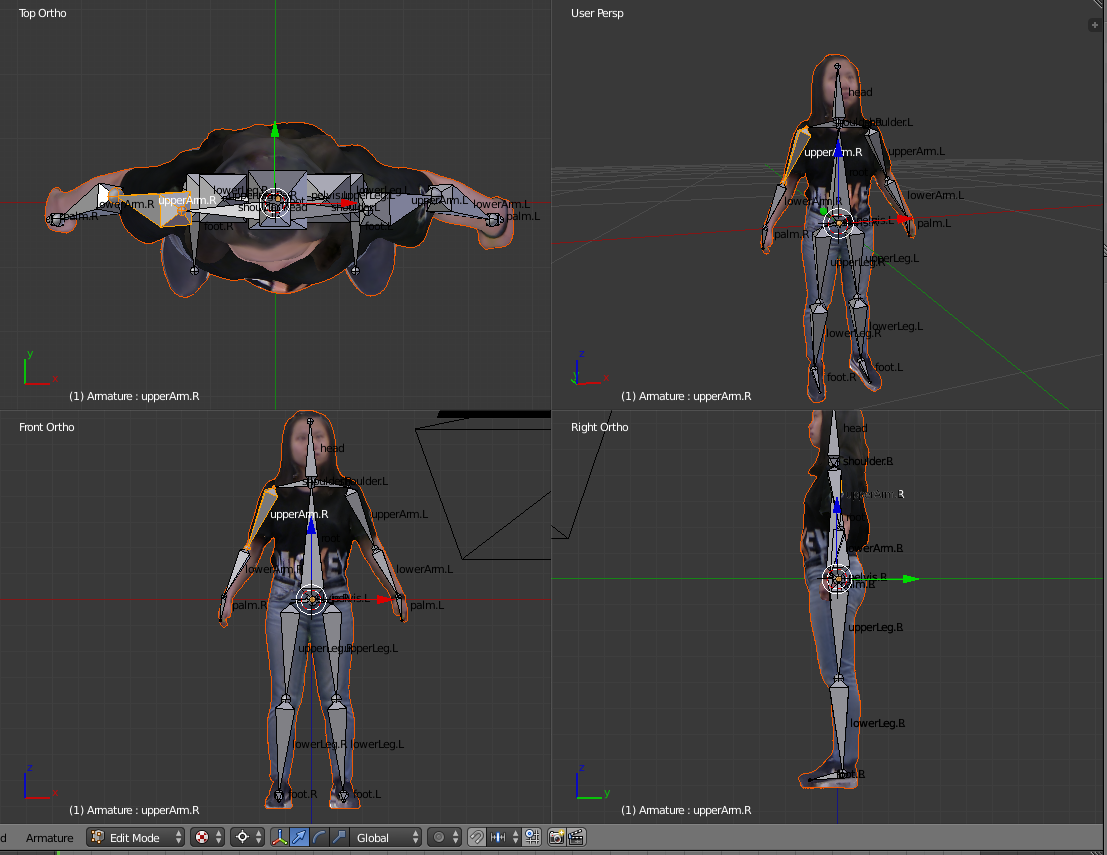}}
 \subfloat[]{\label{fig:riggedPose}\includegraphics[width=0.4\linewidth,{trim=0in 0in 0in 0in,
  clip=true}]{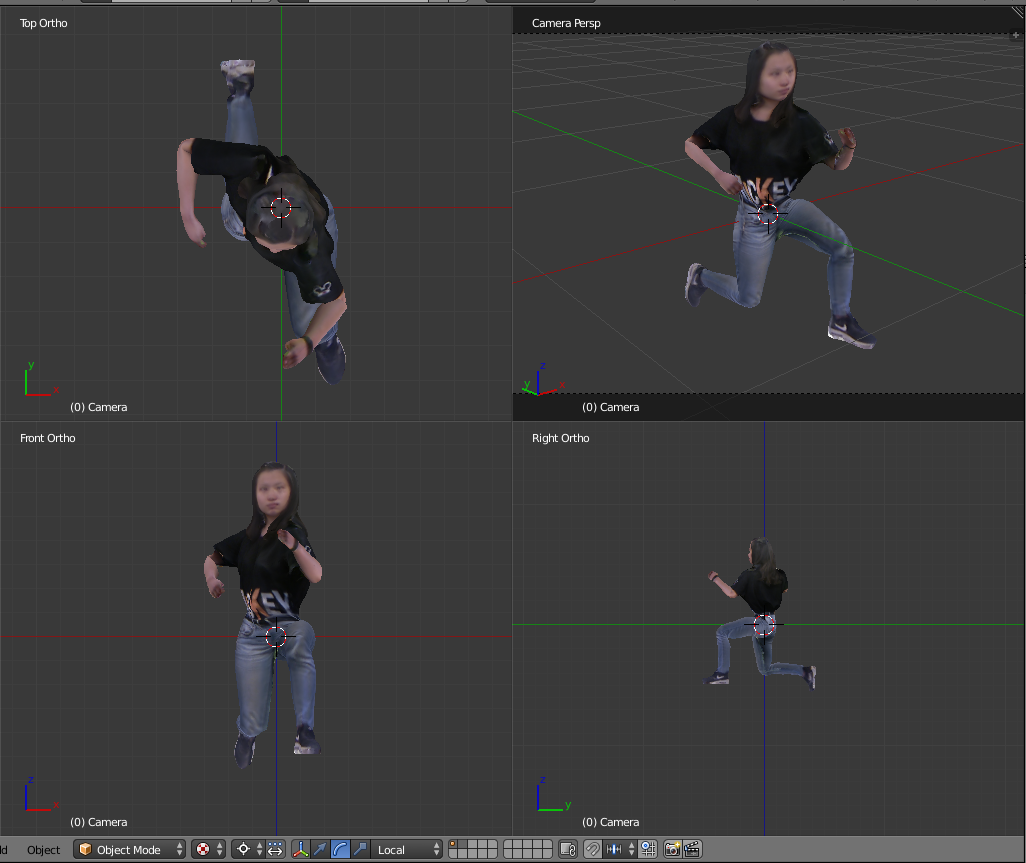}}
    \caption{\fontsize{9}{11}\selectfont Rigged model in Blender software: (a) armature assignment, (b) a generated pose.}
        \vspace{-.1in}
\label{fig:rigged}
    \vspace{-.1in}
\end{figure}
}

\newcommand{\figDatasets}{
\begin{figure}[t]
 \centering
 \subfloat[]{\label{fig:realSet}\includegraphics[width=0.95\linewidth, trim=0in 0in 0in 0in,
  clip=true]{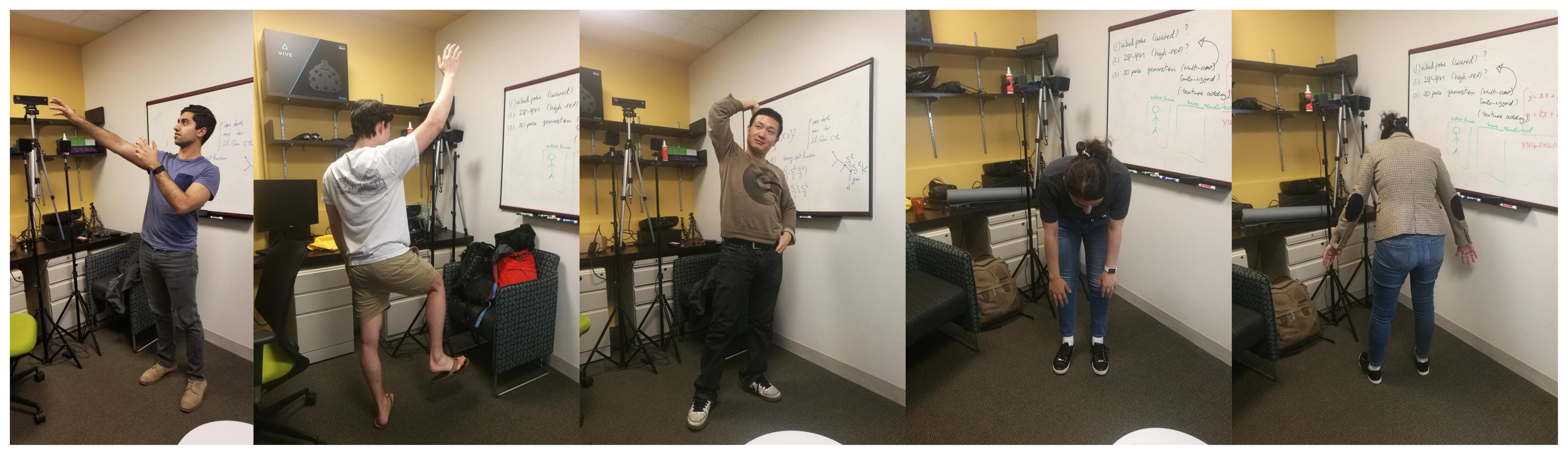}} \\
 \subfloat[]{\label{fig:SYNset}\includegraphics[width=0.95\linewidth,{trim=0in 0in 0in 0in,
  clip=true}]{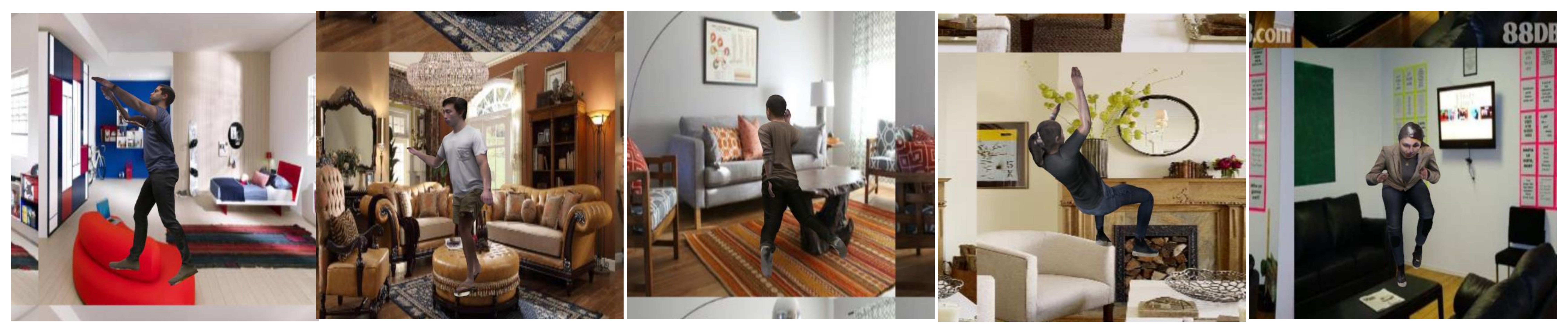}}
    \caption{\fontsize{9}{11}\selectfont (a) Real human pose images, (b) synthesized ScanAva images using our proposed approach.}
            \vspace{-.1in}
\label{fig:datasets}
\end{figure}
}

\newcommand{\figRst}{
\begin{figure}[t]
 \centering
 \subfloat[]{\label{fig:rstGroup}\includegraphics[width=0.45\linewidth, trim=0in 0in 0in 0in,
  clip=true]{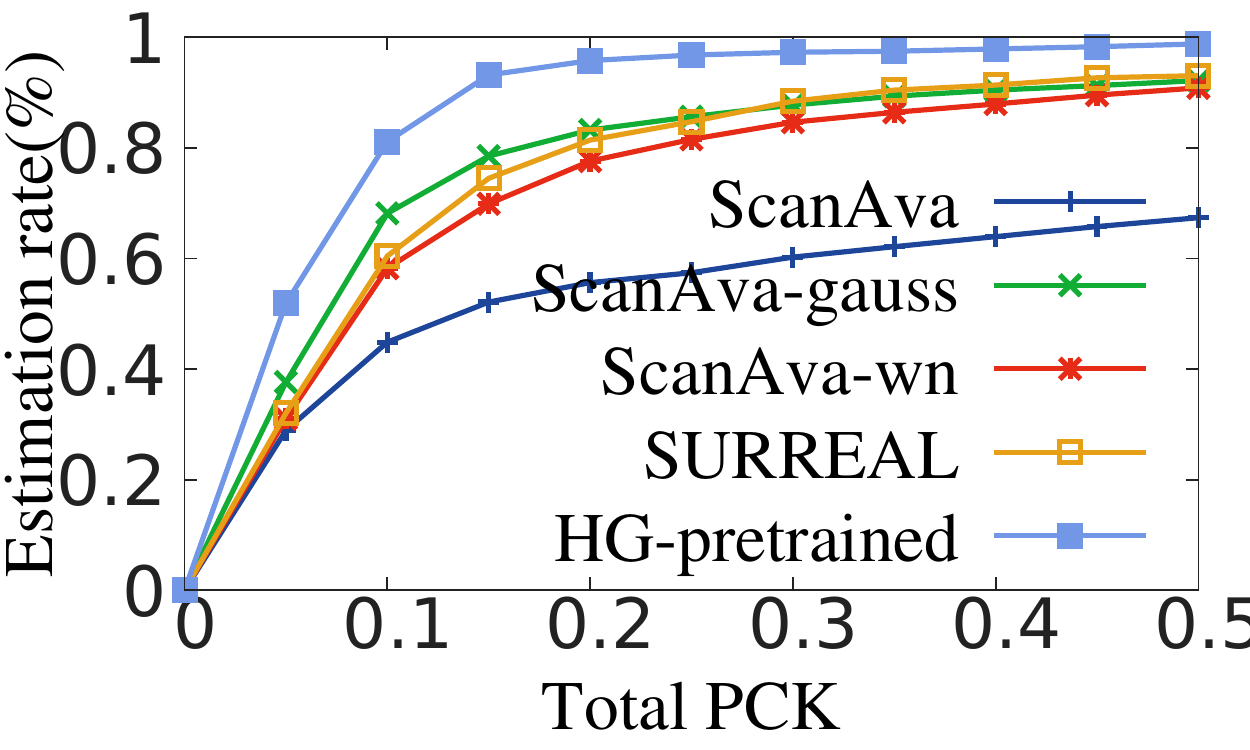}} 
 \subfloat[]{\label{fig:rstInd}\includegraphics[width=0.45\linewidth,{trim=0in 0in 0in 0in,
  clip=true}]{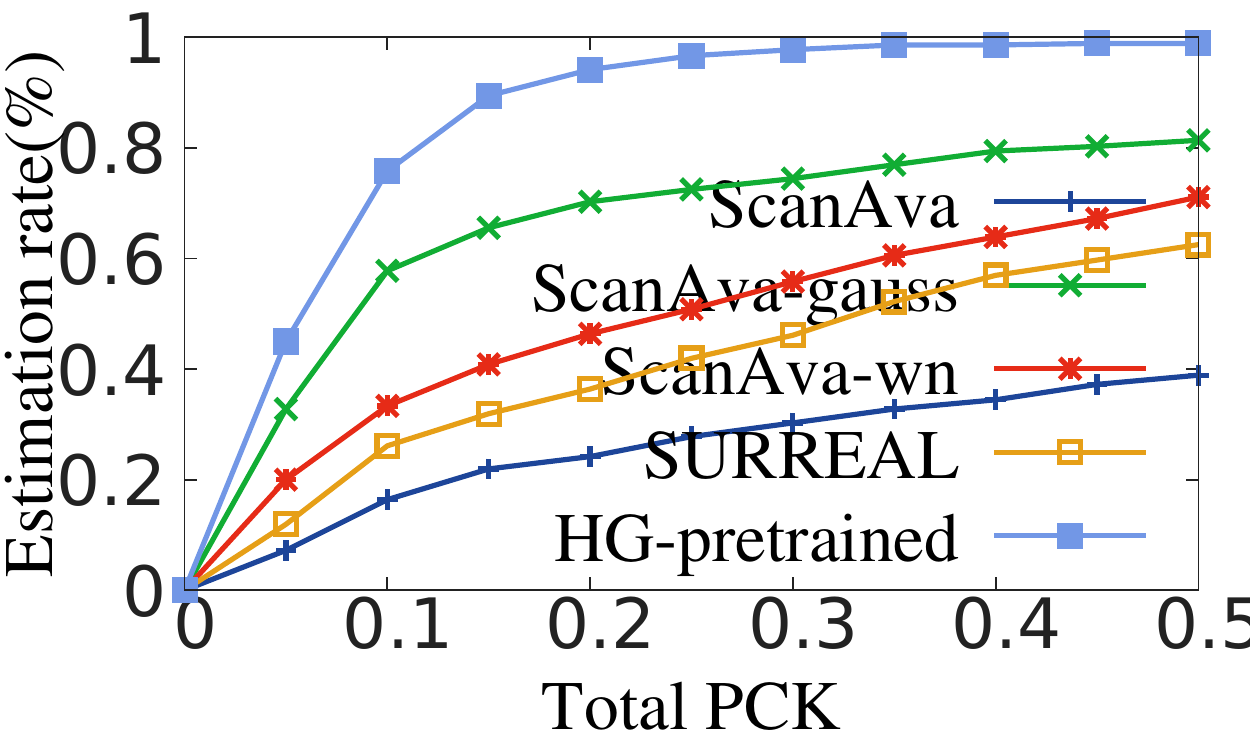}}
    \caption{\fontsize{9}{11}\selectfont Accuracy comparison of a DNN-based pose estimation model trained on different datasets and tested on real human images from: (a) a small group of individuals, (b) a specific person.}
\label{fig:rst}
\end{figure}
}

\newcommand{\figVR}{
\begin{figure}[b]
    \centering
    \includegraphics[width=0.4\textwidth]{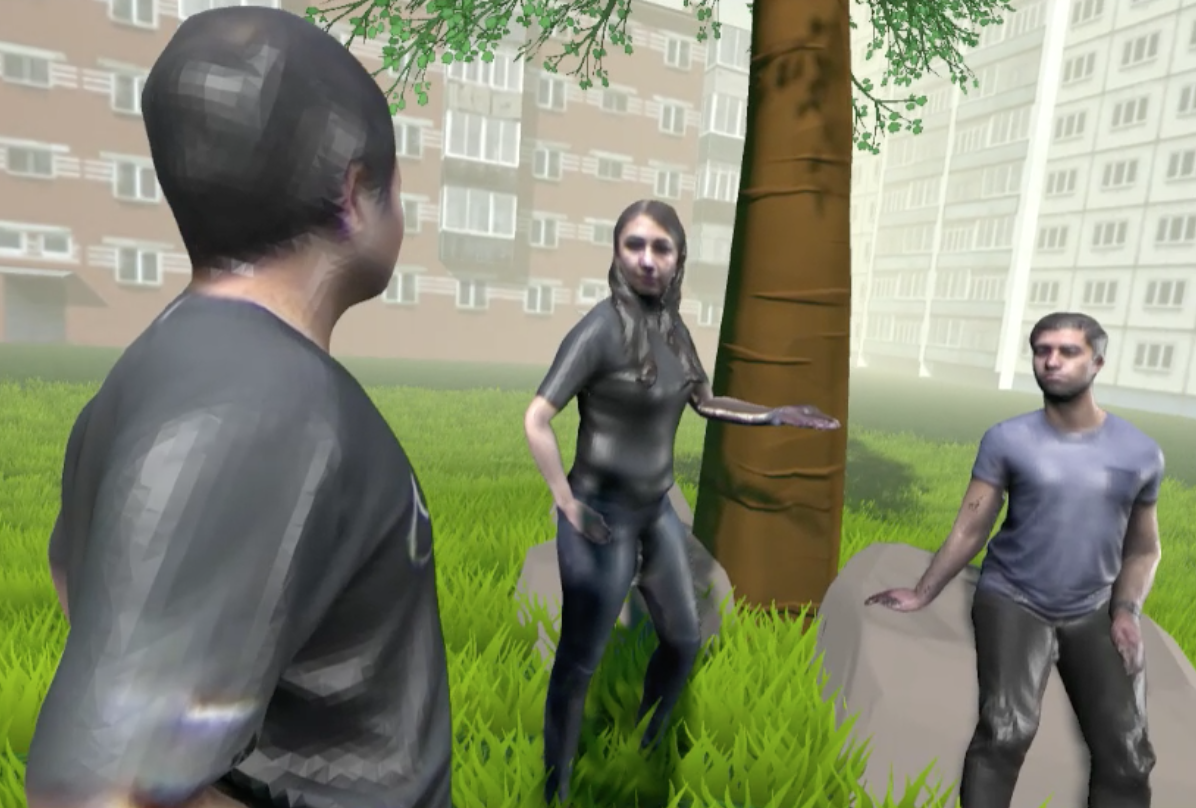}
    \caption{\fontsize{9}{11}\selectfont Avatars of our group members interacting in a VR simulated world.}
    \label{fig:vr}
\end{figure}
}

\newcommand{\figArticulated}{
\begin{figure}[h]
    \centering
    \includegraphics[width=0.4\textwidth]{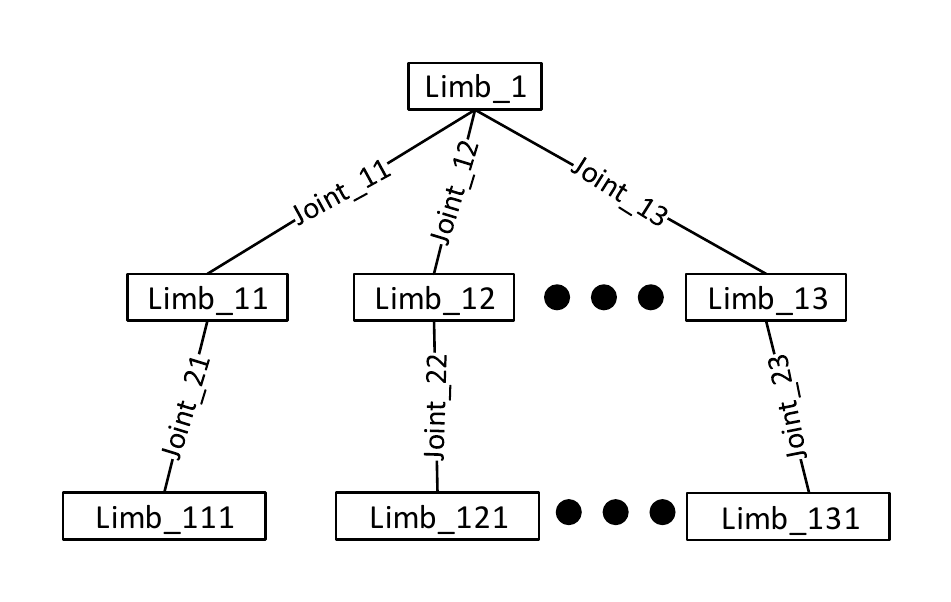}
    \caption{General articulated model shown as graph.}
    \label{fig:articulated}
\end{figure}
}

\newcommand{\figNonMani}{
\begin{figure}[h]
    \centering
    \includegraphics[width=0.4\textwidth]{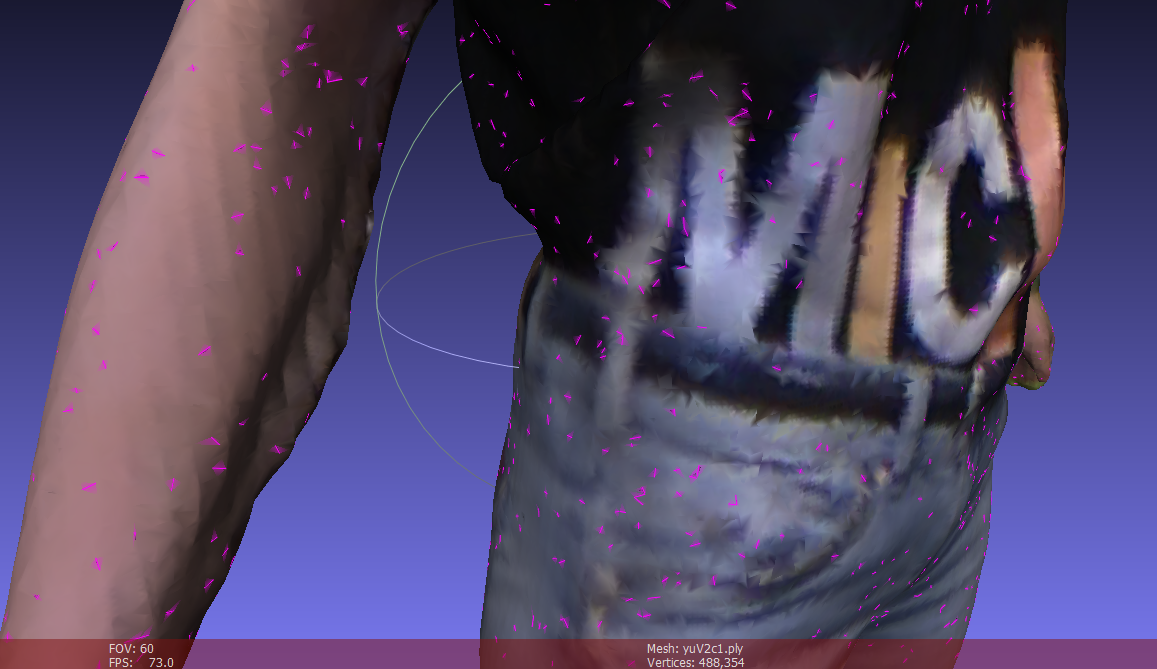}
    \caption{Non-manifold faces after simplification shown in purple.}
    \label{fig:nonMani}
\end{figure}
}

\newcommand{\figSkelCoord}{
\begin{figure}[t]
    \centering
    \includegraphics[width=0.9\textwidth]{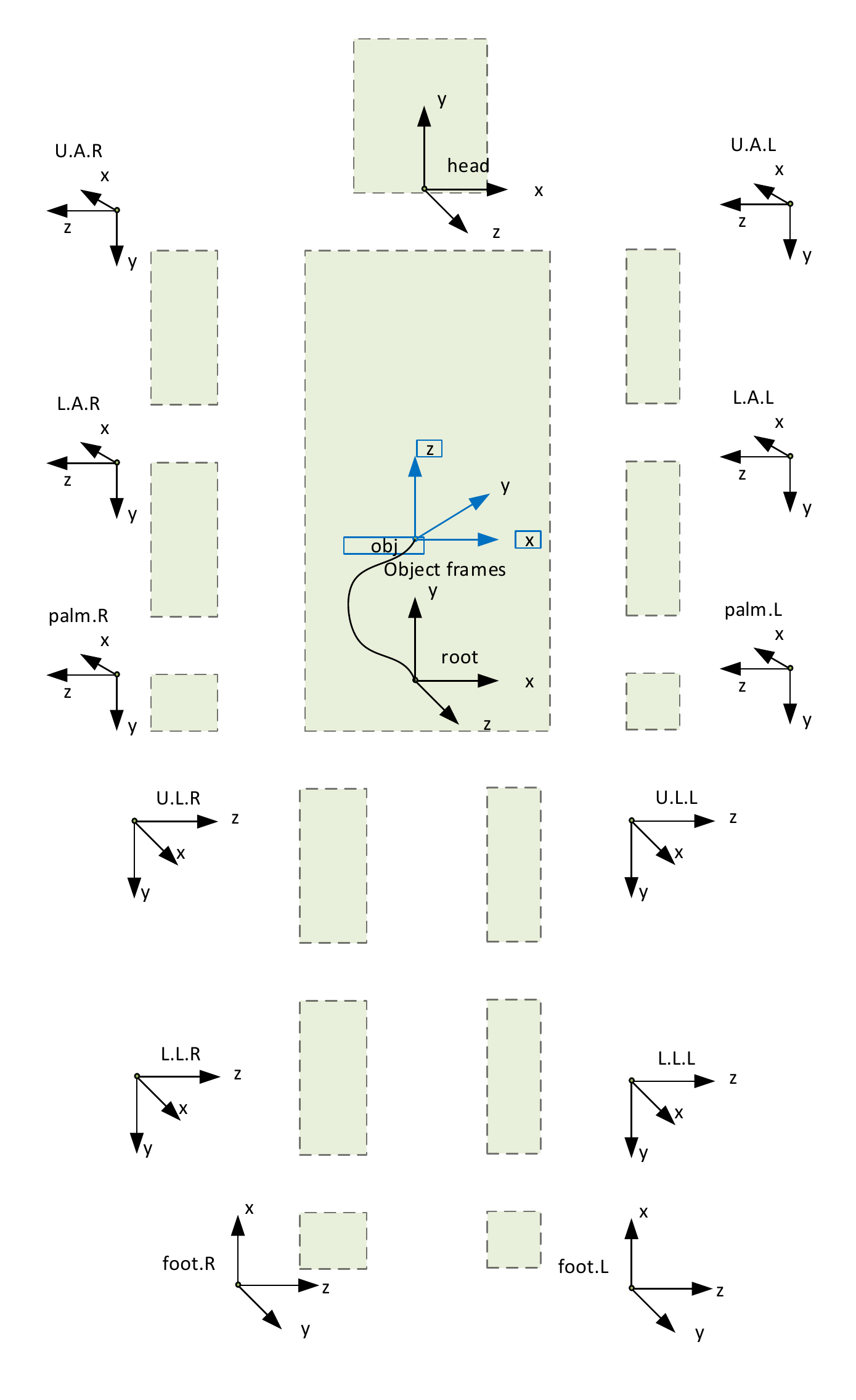}
    \caption{Human skeleton coordinates configuration}
    \label{fig:skelCoord}
\end{figure}
}

\newcommand{\tablresult}{
\begin{table}
\caption{Average breathing rate of restraint voles through different emotional states}
\begin{center}
 \begin{tabular}{c | c | c | c} 
  & Vole \#2 & Vole \#3 & Vole \#4 \\ 
 \hline
 Mass (grams) & 26.5 & 28.5 & 24.45 \\ 
 \hline
 Pre-Odor (bpm) & 216 & 220 & 230 \\
 \hline
 During-Odor (bpm) & 217 & 200 & 202 \\
 \hline
 Post-Odor (bpm) & 228 & 208 & 199 \\

\end{tabular}

\label{tbl:result}
\end{center}
\end{table}
}

\newcommand{\tabOriTable}{
\begin{table}
\caption{Lying orientation encoding table}
\begin{center}
 \begin{tabular}{c | c  c  c c } 
  categories &  $E$ & $W$ & $N$ & $S$ \\ 
 \hline
 $bit_H$ & 1 & 1& 0 & 0 \\ 
 \hline
 $bit_N$ & 1 & 0 & 1 & 0 \\
\end{tabular}
\label{tbl:oriTable}
\end{center}
\end{table}
}

\newcommand{\eqnref}[1]{Eq.~(\ref{eqn:#1})}
\newcommand{\figref}[1]{Fig.~\ref{fig:#1}}
\newcommand{\tblref}[1]{Table~\ref{tbl:#1}}
\newcommand{\secref}[1]{Section~\ref{sec:#1}}
\newcommand{\thmref}[1]{Theorem~\ref{thm:#1}}
\newcommand{\defref}[1]{Definition~\ref{definition:#1}}
\newcommand{\lemref}[1]{Lemma~\ref{lem:#1}}
\newcommand{\com}[1]{\textcolor{red}{#1}}
\maketitle  

% \thispagestyle{empty}

%%%%%%%%% ABSTRACT
\begin{abstract}
Deep learning approaches have been rapidly adopted across a wide range of fields because of their accuracy and flexibility, but require large labeled training datasets. This presents a fundamental problem for applications with limited, expensive, or private data (i.e. small data), such as human pose and behavior  estimation/tracking which could be highly personalized. In this paper, we present a semi-supervised data augmentation approach that can synthesize large scale labeled training datasets using 3D graphical engines based on a physically-valid low dimensional pose descriptor. To evaluate the performance of our synthesized datasets in training deep learning-based models, we generated a large synthetic human pose dataset, called ScanAva using 3D scans of only 7 individuals based on our proposed augmentation approach. A state-of-the-art human pose estimation deep learning model then was trained from scratch using our ScanAva dataset and could achieve the pose estimation accuracy of 91.2\% at PCK0.5 criteria after applying an efficient domain adaptation on the synthetic images, in which its pose estimation accuracy was comparable to the same model trained on large scale pose data from real humans such as MPII dataset and much higher than the model trained on other synthetic human dataset such as SURREAL.

\end{abstract}

%%%%%%%%% BODY TEXT
\section{Introduction}

With the remarkable success of deep neural networks (DNNs) in regression and classification tasks, a significant challenge comes out to be forming a large scale \emph{labeled} datasets to support DNN training requirements. In the computer vision field, widely employed datasets mainly come from collection of real-world images captured from the contexts of interest and labeled through manual processes such as crowdsourcing. A direct benefit of using these datasets for the network training is to preserve authentic information from  real world. However, each image in the training set needs to be labeled for the supervised learning process and it get quite expensive for large datasets \cite{varol2017learning}. On the other hand, for certain applications where data is scarce such as personalized medicine, robot reinforcement learning, environmental/weather behavior prediction, and military applications, forming a large scale dataset itself could be infeasible \cite{liu2017bed}. The million dollar question here is if one can benefit from the flexibility and accuracy of DNNs in small data domains or domains with expensive labeling process by virtually synthesizing large scale labeled datasets. Hence, this paper presents a semi-supervised data augmentation approach that expands the size of a small dataset by synthesizing labeled samples in the physically-valid world contexts, while demonstrating that the trained DNNs using this synthetic dataset are capable of performing a high accuracy estimation task that they are trained for.

Classically, to address the data limitation issue, data augmentation techniques are extensively used especially when it comes to DNN training. Existing data augmentation methods can be seen as a mapping from one domain to itself by linear transformation with random variations, such as scale/orientation augmentation \cite{simonyan2014very}, color augmentation, and random crop per-pixel mean subtraction \cite{krizhevsky2012imagenet,lee2015deeply}, among others. These hand crafted augmentation methods indeed improve the DNN performance in the designated tasks though not significantly \cite{yosinski2014transferable}. They simply ignore the fact that image generation is actually a mapping from the 3D physical world into the 2D image domain, where the camera model is already well defined. The consequence is that classical augmentation methods can only capture superficial variations of the original dataset instead of capturing the semantic meaning of objects in the real world. 

Alternatively, 3D computer-aided design (CAD) models can emulate such geometrical semantic variations in the real world. Majority of the works enabled by the CAD-based data augmentation employ publicly available CAD models \cite{aubry2014seeing,sun2015generating}. 
Some of these models are also templates for specific categories \cite{chen2016synthesizing,varol2017learning}. The extent of data augmentation here is often limited by the existing CAD models, which only provide rough categories or limited by the already existing templates.
Another practical issue is that publicly available CAD models are usually created by human artists and could be in a conceptual ideal condition and lack realistic variations. In contrast, generating (unlabeled) sample images from a large variety of objects, movements, and contexts is fairly achievable in our physical world, in which each sample manifests the physics laws behind our real world. 

In this paper, we merged the benefits of two approaches, (i.e., 3D modeling and (semi)realistic data generation following the physical world laws) and present a data augmentation pipeline for large scale labeled dataset forming that uses the easily collected 3D scans of the target objects (e.g., humans) and move/articulate them in a physically-valid fashion using a 3D graphical engine (e.g., moving human avatars in a virtual environment). 
Although our cost-efficient 3D scans have lower resolution compared to the existing CAD templates and the movements and contexts are virtually synthesized, after a straightforward domain adaption, our approach allows the data augmentation for deep learning purposes in any emerging target objects and can efficiently expand and adjust the movements and contexts based on the application tasks.

%Our approach though less efficient compared to template based synthesizing, our argument comes from its applicability for any emerging target group, the effectiveness to form dataset from limited data for, and scalability to form large dataset for general purpose also.
%traditional dataset forming approaches in computer vision can be easily applied to most of the real world problems as it is straight forward to collect sample images from target population without any constraint. Inspired by this idea, we world like to preserve flexibility of the traditional methods but introduce the automation via 3d engine. As the price of portable scan device goes down, we present a dataset forming pipeline which sample directly from target population with a highly semi-supervised labeling process and large scale augmentation capability. 

\section{Related Work}

When dealing with deep learning in small data domains, fine-tuning already trained DNNs proves to be effective \cite{liu2017bed,bengio2012deep,bengio2011deep,caruana1995learning,yosinski2014transferable}. Fine-tuning is a form of transfer learning, when fine-tuned DNNs applied on the new (but small in size) dataset hugely benefit from knowledge learned from large amount of real world image samples (even being from different domain). However, if the two datasets are very different in nature, fine-tuning would fail since the network is already very fitted to the first dataset and is unable to adopt to the new small dataset unless we substantially increase the size of the second dataset and pay the labeling cost associated with that. These issues inspired us to simulate lots of plausible samples in the context of interest (i.e. the context that only has small dataset available), which allows for training DNNs from scratch rather than just fine-tuning them.

%Deep learning though effective, require large data to support the training process. In case of limited data, fine tuning process prove to be effective \cite{bengio2012deep,bengio2011deep,caruana1995learning,yosinski2014transferable} but few to explore the way to train DNN from scratch when data is limited. As we believe fine tuning itself is actually benefit from knowledge learned from large amount of real world samples, we seek to machine's self-inspiration by simulating lots of plausible samples when given limited information.

\subsection{Classical Dataset Forming}
A common way to form datasets in computer vision field is collecting real images directly and manually label them. Most influential datasets are formed in this way including ImageNet \cite{krizhevsky2012imagenet} for object classification, Cityscapes \cite{Cordts2016Cityscapes} for scene segmentation, LSUN \cite{yu15lsun} for scene understanding, and MPII human pose \cite{andriluka14cvpr} and LSP \cite{Johnson10} datasets for human pose estimation. These datasets preserve the real world information authentically and are most effective to train DNN models for practical applications 
\cite{zhou2017scene,everingham2010pascal,zhou2017scene}. 
Data augmentation usually comes during training session, which usually include augmentation in scale, color, shift, or mirror, which is limited to superficial variations of the image \cite{simonyan2014very}.

\subsection{Synthetic Dataset Forming}
Synthetic data has already been employed to form large datasets and provides convenience to control the generation process with exact parameters \cite{sun2015generating}. In early works, synthetic data was mainly employed to provide additional information to facilitate the detection/estimation process. For example, in \cite{liebelt2010multi}, the geometric information from 3D CAD models is combined with the real image appearance to improve object detection and pose estimation for bicycles and cars. Shape models and also the probabilistic models are also learned from CAD models \cite{stark2010back,sun2009multi}. Another benefit of synthesizing data is the possibility to automatically generate enough labeled data for supervised learning purposes \cite{su2015render}. Authors in  \cite{dosovitskiy2015flownet} studied an optical flow estimation algorithm based on synthesized images of a 3D moving chair. Virtual KITTI dataset with synthesized car videos is also employed to train multi-object tracking algorithms \cite{gaidon2016virtual}. 

More complicated articulated 3D models are also studied, among which the human body draws most attention due to the extensive applications associated with studying human pose, gestures, and activities. Synthesized human data has been employed for 2D/3D pose estimation \cite{pishchulin2012articulated,qiu2016generating,romero2015flowcap,chen2016synthesizing,du2016marker,ghezelghieh2016learning,okada2008relevant} 
and pedestrian detection \cite{marin2010learning,pishchulin2012articulated}. No matter the synthesized human data is collected from publicly available graphical 3D models or from generalized templates, they can hardly represent every individual in various contexts. Even a morphable human template such as SCAPE method (Shape Completion and Animation for PEople) can hardly represent a person in different clothing conditions \cite{anguelov2005scape}.

\subsection{Our Contributions}
With ever lowering price of portable 3D scanning devices, in this paper, we present an alternative way for synthetic dataset forming which combines the flexibility of classical data collection method and the automation of 3D engines as shown in \figref{physicsBased}. The main contributions of this work include: (1) providing a rapid and cost-efficient pipeline to form large scale labeled datasets from a target set of objects in various contexts; (2) exploring the way to minimize domain shift between source (virtual) and target (real) correspondence with limited data from the target domain; (3) demonstrating the proposed approach on human pose estimation problem by training a state-of-the-art DNN model with the limited 3D human scan samples from scratch and evaluating the trained DNN pose estimation performance on real world human images with comparison with the leading real and synthetic datasets; and (4) publicly releasing our synthetic human pose datasets called scanned avatars or ``ScanAva'' and our dataset generation tools in the Augmented Cognition Lab (ACLab) webpage. \footnote{This paper has dataset available at \href{http://www.coe.neu.edu/Research/AClab/ScanAva/ScanAva.zip}{ScanAva} and the code at \href{https://github.com/ostadabbas/ScanAvaGenerationToolkit}{GitHub} provided by the authors. Contact the corresponding author for further questions about this work.}% <-this % stops a space

\figPhysicsBased

\section{A Semi-Supervised Data Augmentation Approach}

%Famous large datasets, though successful, we can hardly say it fairly covers the universal domain. 

No matter if it is collected from objects, animals, humans, or scenes, considering the various appearances, pose states, and their combinations, a dataset even very large one can hardly cover the whole space of its universal attribute domain (i.e., feature space).
%and could be deemed as a manifold inside that high dimensional space. 
%\com{why feature space get sparse if it covers borad categories? lower performance? but in this way with large dataset, they generalize well}
When the categories covered by the dataset are too broad, its feature space can get easily sparse causing low performance learning due to the highly probable over-fitting problem. In contrast, within a specific and well-defined category, if data with enough granularity is available, it could form a low dimensional manifold within the feature space leading to a better (manifold) learning especially when DNNs are used.
%Admittedly, they generalize well during training because they have large quantity of samples in each categories with low granularity. But in case that we would like to further classify within categories, it needs further labeling and can hardly keep the performance because of lack of data in finer categories. 
Due to the high cost of collecting and labeling highly granular dataset, we present a pipeline to form a large labeled dataset with controllable granularity. In particular, we synthesized large scale datasets from a highly articulated object, ``human body'' and validated the dataset quality by performing human pose estimation using DNN models trained from scratch purely on these synthesized datasets, called ScanAva. 

Let's assume image $I= f(G, \boldsymbol{\theta}, E_v)$ contains a human figure, where $G$ is the person's geometry appearance, $\boldsymbol{\theta}$ is the person's pose information, and $E_v$ is the environment and background's parameters. Our dataset ScanAva forming pipeline then includes (see \figref{physicsBased}): (1) collecting the appearance model $G$ by an affordable 3D scanning process; (2) rigging and one-time limb labeling of the 3D scans (i.e., avatars) based on their articulation for valid human reposing; (3) defining a low dimensional pose descriptor for physically-valid reposing; (4) 3D data augmentation by changing the pose information $\boldsymbol{\theta}$ with controllable granularity based on a given application; and (5) rendering 2D images from the 3D data with different environment's parameters $E_v$. These steps lead to generation of our ScanAva datasets, in which each image has a human figure with the person's pose physically-valid and precisely labeled.

\subsection{Geometry Appearance Acquisition via 3D Scanning} \label{sec:PMD}
% We employ a physical model structure $M_p$ to contain all the attributes of the target object. The most important attribute is
An object's geometry appearance $G$ is a major component that affects its image. We  employed the conventional 3D model formats such as 3DS models (*.3ds), Wavefront OBJ (*.obj), and PLY (*.ply) to represent $G$. In articulated cases, $G$ contains skeleton with multiple entries to represent each moving part $G(i)$, where $i$ stands for the part index. In a rigid body case, $G$ simply reduce to one component in our model. In an articulated human body, based on its biomechanics and skeleton, we predefined 14 moving parts (i.e., limbs) as head (H), torso (T), left upper arm (LUA), left lower arm (LLA), left palm (LP), right upper arm (RUA), right lower arm (RLA), right palm (RP), left upper leg (LUL), left lower leg (LLL), left foot (LF), right upper leg (RUL), right lower leg (RLL), and right foot (RF), as shown in \figref{skeleton}. The limbs are articulated together with joints. Each joint state is described by a rotation angle. We employed a state vector $\boldsymbol{\theta}=\{\theta_1, \theta_2,..., \theta_n\}$ to describe the pose information. Our model can be described by a graph where limb geometry acts as node and state vector describes the edges between limbs. This graph varies depending on the target configuration. 

%The geometry models can be attained by existing 3D scanning software such as Agisoft \cite{Agisoft}. A general version of articulated model is shown in \figref{articulated}.
%\figArticulated

\figScan

To get the 3D geometry model, we employ a Microsoft Kinect v1 to  perform 3D scanning of the human body using off-the-shelf components and software. Subject stands on a automatic rotator in front of the Kinect sensor. We employed a commercially available software, Skanect to extract 3D information from the scanning frames \cite{Skanect}. The scanning process is shown on the monitor to give realtime feedback. Our 3D scanning setup is shown in \figref{scanProcess}. When the space is limited, the camera's  field of view cannot cover the whole body. In this case, we will pitch the camera up and down to extend the sensing area. The whole body scan can be achieved by stitching them together.

\subsection{Rigging for Reposing}
Rig is essentially a digital skeleton bound to a 3D mesh that consists of joints and bones. Joints and bones in the rig can be moved by altering the pose state descriptor $\boldsymbol{\theta}$, which leads to reposing and animation of the 3D models. To rig the 3D scanned model (avatars), we employed an existing animation toolkit "Blender" to manipulate the body parts and give them the pose we wish. An armature is assigned to the scanned model. The root bone is set at the center of the pelvis. Each arm has an upper arm, lower arm and palm bone. Each leg has an upper leg, a lower leg and a foot bone. The head bone is also assigned to it. A demo of rigged model is shown in \figref{riggedModel}. From the rigged model, we can easily manipulate the avatar to generate the pose we need by manipulating the pose state vector, $\boldsymbol{\theta}$. An example generated pose is shown in \figref{riggedPose}. The corresponding 2D image is achieved by re-projection of the 3D model into the image domain. In cases where scanned raw models have defects such as holes, over complicated details, or non-manifold geometries, we can optionally employ Meshlab open-source software or equivalent toolkit for preprocessing including simplification, filling holes, and also non-manifold geometries removal \cite{MeshLab}. 
\figRigged

\subsection{Manifold Pose Generation via a Low Dimensional Pose Descriptor} \label{sec:manifold}

To give a specific example, we model the human pose as follows. Following humanoid robot convention \cite{kajita2014introduction}, we define shoulder, neck, and hip as spherical joints, elbow and knee as revolute joints and wrist and ankle as universal joints. Higher degree joints can be decomposed into multiple one-degree joints. Each arm/leg has 6 degree of freedom (DOF) and with a 3 DOF neck, giving $\boldsymbol{\theta}$ a dimensionality of $n=27$. Pose space is actually a constrained manifold: not all 27-dimensional vectors represent valid poses. 
We assume two ways to generate valid descriptor. One way is following kinematic constrains to make generated descriptor physically valid which is helpful when motion data is limited. 
The other is direct sampling from data lies on such manifold such as motion capture data \cite{mocap}. 
In first method, the two constraints considered during the generation phase are joint angle constraints and global orientation constraints. We will use a constraint matrix to indicate the range of each state variable as $[\theta_{is},\theta_{ie}]$ for $i\in [1:n]$, where $\theta_{is}$ and $\theta_{ie}$ stand for the low and high acceptable ranges of the state $\theta_i$. For example, for the human elbow joint, the possible rotation range is around from 0$^{\circ}$ to 145$^{\circ}$. In addition, depending on the application, there might be global orientation constraints. For example, for in-bed poses, the torso will lie approximately parallel to the bed. The Euler description for the body orientation as $(\alpha,\beta,\gamma)$ will show the relative orientation of the body with respect to the world frame \cite{craig2005introduction}. In the context of walking, we can simply limit the Euler angles to a range to mimic the up straight poses, for example $\alpha, \gamma \in [-30^{\circ},30^{\circ}]$. Therefore, both joint angle and global orientation constraint types can be modeled using range bounds. Within these bounds, poses can be generated from a uniform random distribution, or they can be generated procedurally using a grid-based approach. 

Since this is used for training a DNN, we use the random approach to take advantage of a common training optimization, Stochastic Gradient Decent (SGD) \cite{bottou2010large}. In SGD, a fixed-size batch is randomly selected from fixed size dataset with random variation such as crop and scaling. In our work, random generation is equivalent to random selection from an infinite training set as we sample from a virtually continuous pose manifold. 

% \subsubsection{Pose Filtering and Editing}  \label{sec:PF}
% The simple joint constraints used during the generation phase can still allow for impossible poses, such as poses with self-intersection and poses not in static equilibrium. The low-dimensionality of the pose subspace makes it efficient to generate poses and filter for feasibility. The intersection constraint filter employs collision detection mechanisms available in commercial game engines, such as Unity \cite{Unity}. Poses that violate self-intersection constraints are simply eliminated. 

\subsection{Rendering 2D Images with Different Environment's Parameters}

Besides the subject state, we also introduce environment's parameters $E_v$ to render realistic images. The environment includes all items in the scene and also the lighting and camera's parameters which can be simply described. Since in the human pose estimation problem, we mainly care about the person in the scene, we have fixed the camera parameters to 35mm focal length and simplified this description by camera view point under spherical coordinate. For the background, instead of parameterized description, we directly sample from a context image dataset such as LSUN \cite{yu15lsun} to generate images with different backgrounds. 

One direct benefit of our approach is that we can describe the synthesized dataset in a more compact way. In our running example of human body, we only need a rigged model (with 300,000 face mesh with the size around 35MB) and a low dimension descriptors including $\boldsymbol{\theta}$ and $E_v$, in which for 2000 pose information is under 1MB compared to a standard dataset which with this many 512$\times$512 pose images can get to up to 2GB in size. Another benefit is that classical augmentation methods like shift and crop operations can be simply simulated by changing the relative position of the human with respect to the camera. Therefore, our approach can also accommodate these augmentation methods besides the physically-valid state variable augmentations.

\section{Synthetic Dataset Quality Evaluation}

We evaluated the quality of our synthesized datasets by testing the human pose estimation network's performance when trained on ScanAva datasets from scratch. To generate different versions of the ScanAva datasets, we collected 3D scans from 7 participants, in which 4 of them repeated the scanning procedure with various clothes. We generated 2000 images for each 3D scan with random pose selected from CMU MoCap dataset \cite{mocap} and random background from indoor environment of LSUN\cite{yu15lsun}. We formed totally 15 ScanAva datasets to evaluate their quality for pose estimation DNN training when tested on: (1) a small group of individuals and (2) one specific person in different clothes.
% We also collect same dataset with random pose uniformly sampled with only kinematic constrain to test its performance with limited information collected from real world. 
From each participant, 10 to 15 corresponding 2D images are also captured using an iPhone 7 camera to be used as the real world test dataset, in which individuals were asked to give random poses as they wish. Several demo images are shown in \figref{datasets}. 

\figDatasets

\subsection{Synthetic vs. Real Domain Adaptation}

Even when data is collected  from the exact same person, domain shift is a common issue between synthetic and real world/human images. It is also known even real world images collected with different devices are affected by this issue \cite{gong2012geodesic}. To minimize domain shift effects in learning and estimation, people try to make both domains as similar as possible, for example aligning the 2nd order statistics of the training and test datasets \cite{sun2016correlation}. Visually perceiving the synthetic and real images of a given person revealed that although the profiles are quite similar in both domains, the details are different. Therefore, since we aim at a quick and efficient large scale dataset forming, we applied two direct modifications to weaken such differences in details by applying (1) Gaussian filtering (ScanAva-gauss) and (2) direct white noise (ScanAva-wn) on images in both domains to make their appearances as similar as possible.

\subsection{Pose Estimation Performance of Trained DNN Models}

To evaluate the quality of the synthesized datasets, we employed a state-of-the-art DNN-based 2D human pose estimation algorithm, a stacked hourglass model \cite{newell2016stacked} and train it from scratch with our synthesized ScanAva datasets and compared their pose estimation performance with stacked hourglass models trained on real human pose image dataset, MPII \cite{andriluka14cvpr} (HG-pretrained) and synthetic human pose dataset, SURREAL \cite{varol2017learning}. During the training procedure, we kept the hyper-parameters of the stacked hourglass model the same between experiments to have a fair comparison among different training datasets. The chosen hyper-parameters were learning rate 2.5e-4, 30 epochs, 8000 iterations, and 8 stacked netwroks. For the pose estimation performance evaluation, we employed the conventional pose estimation metric, the probability of corrected keypoints (PCK) standard \cite{Johnson10,wei2016convolutional} to test the estimated joint locations against the ground truth locations on real human pose images. 

In the first experiment, we synthesized the ScanAva datasets using 7 participant 3D scan data without (ScanAva-no) and with domain adaption (ScanAva-gauss and ScanAva-wn). The pose estimation accuracy results comparing the performance of the stacked hourglass models trained on these ScanAva datasets as well as models trained on MPII (called HG-pretrained) and SURREAL datasets and tested on our real human test dataset is shown in \figref{rstGroup}. From the figure, it is clear that pose estimation DNN model trained directly on raw synthetic dataset shows poor estimation performance and domain adaptation by applying Gaussian filtering or even adding white noise improves the model performance significantly. Surprisingly, in high standard criteria like PCK0.2, the model trained on ScanAva-gauss even surpasses the one trained on SURREAL dataset, which in fact  includes thousands of appearance variations compared to our limited subject dataset. Nonetheless, there is an obvious gap between the performance of the models trained on synthetic data vs. real data, as one expects. 

%But we should be aware that the dataset is generated in a time and cost efficient way under semi-supervising. 

In the second experiment, to test the capability of our dataset for individualized pose estimation training, we synthesized 11 datasets of one participant with varying clothes and also collected corresponding real images as a test dataset. To fairly evaluate the generalization ability of the pose estimation model based on these datasets, we trained the stacked hourglass DNN with scans from only 9 clothes and left the rest of scans out of model training. According to the results of the domain adaptation from \figref{rstGroup}, we used Gaussian filter as optimal domain adapter in this experiment. The pose estimation results are shown in \figref{rstInd}, where although models trained on our datasets falls behind the HG-pretrained model, but when domain adapted surpass the model trained on SURREAL dataset with a big margin. We believe performance drop of the DNN trained on SURREAL mainly comes from the incompleteness of its templates, which only contains the bare human body shapes instead of clothed ones, while in the real world pose detection problems, people are rarely naked and come in the variety of clothes. These outcomes emphasize that for person-specific pose estimation/tracking in applications such as gaming, human-computer interaction, and daily activity monitoring our approach can quickly and efficiently build a large scale labeled dataset to be used for training of robust and accurate DNN models.

%This is possibly because there is bulky clothed one in test set which SURREAL template obviously doesn't take into account. This is also a evidence of our proposal of high diversity in human poses and appearance which can hardly be generalized in a single template or dataset. But for a specific application of a narrowed category, our method can quickly generate large plausible samples to approximate the target manifold. 

\figRst

% 1. Time cost of different labeling. 

% 2. Accuracy within virtual domain.  With transfer domain for real images. 

% It reaches nearly 99\% in virtual domain, so we only care about its performance on real images. 

% virtual on real with different domain adaptation method. 
% 1. P7 accuracy with different pre-processing method. gaussFt, noise. 
% gaussFt gives the best result. employs gaussFt method on following test. 
% 2. different angle cross test. 
% Parameterized dataset concept. Under specific context, pinpoint the specific parameter can help to improve the performance. refer the in-bed pose estimation. 

% 3. individualized test. performance drop,  but still around 80\%. 
% Limited appearance variation. But it has the potential to be extended for individualized pose estimation. 

\figVR

\section{Discussion on Future Work}
In this paper, we presented a fast and cost-efficient pipeline to form large scale labeled datasets from a small numbers of available samples via a semi-supervised synthetic data generation approach. In an exploration for a time-efficient domain adaptation method, without even having access to the target domain data, we achieved significant performance improvement using Gaussian filtering which made the synthetic and real data very similar in their appearance. Though our dataset forming approach is only tested on pose data from a small group of individuals, it can be seen as an alternative way of data collection for any general purpose. For example, if we scale up the numbers of easily collected 3D scans, it can possibly show a reasonable performance for general human pose estimation. The proposed pipeline is not limited to the applications to forming large labeled dataset for DNN training, as it also provides the utility to generate personalized avatars. A demo of several 3D scanned model of our group members are shown in a virtual reality (VR) environment in \figref{vr}.

\section{Acknowledgement}
This research was supported by NSF grant \#1755695. The authors would also like to thank Naveen Sehgal who actively participated in the ScanAva dataset formation procedure at the Augmented Cognition Lab (ACLab) in the Electrical and Computer Engineering Department at Northeastern University.

%\clearpage

\bibliographystyle{splncs}
\bibliography{paper}

\begin{thebibliography}{10}
\providecommand{\url}[1]{\texttt{#1}}
\providecommand{\urlprefix}{URL }
\providecommand{\doi}[1]{https://doi.org/#1}

\bibitem{MeshLab}
Meshlab. \url{http://www.meshlab.net/}, accessed: 2018

\bibitem{mocap}
Cmu graphics lab motion capture database. http://mocap.cs.cmu.edu/  (2018)

\bibitem{Skanect}
{Skanect 3D Scanning Software By Occipital}.
  \url{http://skanect.occipital.com/} (Accessed: 2018)

\bibitem{andriluka14cvpr}
Andriluka, M., Pishchulin, L., Gehler, P., Schiele, B.: 2d human pose
  estimation: New benchmark and state of the art analysis. IEEE Conference on
  Computer Vision and Pattern Recognition (CVPR)  (June 2014)

\bibitem{anguelov2005scape}
Anguelov, D., Srinivasan, P., Koller, D., Thrun, S., Rodgers, J., Davis, J.:
  Scape: shape completion and animation of people. ACM transactions on graphics
   \textbf{24}(3),  408--416 (2005)

\bibitem{aubry2014seeing}
Aubry, M., Maturana, D., Efros, A.A., Russell, B.C., Sivic, J.: Seeing 3d
  chairs: exemplar part-based 2d-3d alignment using a large dataset of cad
  models. Proceedings of the IEEE conference on computer vision and pattern
  recognition pp. 3762--3769 (2014)

\bibitem{bengio2012deep}
Bengio, Y.: Deep learning of representations for unsupervised and transfer
  learning. Proceedings of ICML Workshop on Unsupervised and Transfer Learning
  pp. 17--36 (2012)

\bibitem{bengio2011deep}
Bengio, Y., Bergeron, A., Boulanger-Lewandowski, N., Breuel, T., Chherawala,
  Y., Cisse, M., Erhan, D., Eustache, J., Glorot, X., Muller, X., et~al.: Deep
  learners benefit more from out-of-distribution examples. Proceedings of the
  Fourteenth International Conference on Artificial Intelligence and Statistics
  pp. 164--172 (2011)

\bibitem{bottou2010large}
Bottou, L.: Large-scale machine learning with stochastic gradient descent.
  Proceedings of COMPSTAT'2010 pp. 177--186 (2010)

\bibitem{caruana1995learning}
Caruana, R.: Learning many related tasks at the same time with backpropagation.
  Advances in neural information processing systems pp. 657--664 (1995)

\bibitem{chen2016synthesizing}
Chen, W., Wang, H., Li, Y., Su, H., Wang, Z., Tu, C., Lischinski, D., Cohen-Or,
  D., Chen, B.: Synthesizing training images for boosting human 3d pose
  estimation. 3D Vision (3DV), 2016 Fourth International Conference on pp.
  479--488 (2016)

\bibitem{Cordts2016Cityscapes}
Cordts, M., Omran, M., Ramos, S., Rehfeld, T., Enzweiler, M., Benenson, R.,
  Franke, U., Roth, S., Schiele, B.: The cityscapes dataset for semantic urban
  scene understanding. Proc. of the IEEE Conference on Computer Vision and
  Pattern Recognition (CVPR)  (2016)

\bibitem{craig2005introduction}
Craig, J.J.: Introduction to robotics: mechanics and control, vol.~3. Pearson
  Prentice Hall Upper Saddle River (2005)

\bibitem{dosovitskiy2015flownet}
Dosovitskiy, A., Fischer, P., Ilg, E., Hausser, P., Hazirbas, C., Golkov, V.,
  van~der Smagt, P., Cremers, D., Brox, T.: Flownet: Learning optical flow with
  convolutional networks. IEEE International Conference on Computer Vision pp.
  2758--2766 (2015)

\bibitem{du2016marker}
Du, Y., Wong, Y., Liu, Y., Han, F., Gui, Y., Wang, Z., Kankanhalli, M., Geng,
  W.: Marker-less 3d human motion capture with monocular image sequence and
  height-maps. European Conference on Computer Vision pp. 20--36 (2016)

\bibitem{everingham2010pascal}
Everingham, M., Van~Gool, L., Williams, C.K., Winn, J., Zisserman, A.: The
  pascal visual object classes (voc) challenge. International journal of
  computer vision  \textbf{88}(2),  303--338 (2010)

\bibitem{gaidon2016virtual}
Gaidon, A., Wang, Q., Cabon, Y., Vig, E.: Virtual worlds as proxy for
  multi-object tracking analysis. arXiv preprint arXiv:1605.06457  (2016)

\bibitem{ghezelghieh2016learning}
Ghezelghieh, M.F., Kasturi, R., Sarkar, S.: Learning camera viewpoint using cnn
  to improve 3d body pose estimation. 3D Vision (3DV), 2016 Fourth
  International Conference on pp. 685--693 (2016)

\bibitem{gong2012geodesic}
Gong, B., Shi, Y., Sha, F., Grauman, K.: Geodesic flow kernel for unsupervised
  domain adaptation. Computer Vision and Pattern Recognition (CVPR), 2012 IEEE
  Conference on pp. 2066--2073 (2012)

\bibitem{Johnson10}
Johnson, S., Everingham, M.: Clustered pose and nonlinear appearance models for
  human pose estimation. Proceedings of the British Machine Vision Conference
  (2010), doi:10.5244/C.24.12

\bibitem{kajita2014introduction}
Kajita, S., Hirukawa, H., Harada, K., Yokoi, K.: Introduction to humanoid
  robotics, vol.~101. Springer (2014)

\bibitem{krizhevsky2012imagenet}
Krizhevsky, A., Sutskever, I., Hinton, G.E.: Imagenet classification with deep
  convolutional neural networks. Advances in neural information processing
  systems pp. 1097--1105 (2012)

\bibitem{lee2015deeply}
Lee, C.Y., Xie, S., Gallagher, P., Zhang, Z., Tu, Z.: Deeply-supervised nets.
  Artificial Intelligence and Statistics pp. 562--570 (2015)

\bibitem{liebelt2010multi}
Liebelt, J., Schmid, C.: Multi-view object class detection with a 3d geometric
  model. Computer Vision and Pattern Recognition (CVPR), 2010 IEEE Conference
  on pp. 1688--1695 (2010)

\bibitem{liu2017bed}
Liu, S., Yin, Y., Ostadabbas, S.: In-bed pose estimation: Deep learning with
  shallow dataset. arXiv preprint arXiv:1711.01005  (2018)

\bibitem{marin2010learning}
Marin, J., V{\'a}zquez, D., Ger{\'o}nimo, D., L{\'o}pez, A.M.: Learning
  appearance in virtual scenarios for pedestrian detection. Computer Vision and
  Pattern Recognition (CVPR), 2010 IEEE Conference on pp. 137--144 (2010)

\bibitem{newell2016stacked}
Newell, A., Yang, K., Deng, J.: Stacked hourglass networks for human pose
  estimation. European Conference on Computer Vision pp. 483--499 (2016)

\bibitem{okada2008relevant}
Okada, R., Soatto, S.: Relevant feature selection for human pose estimation and
  localization in cluttered images. European Conference on Computer Vision pp.
  434--445 (2008)

\bibitem{pishchulin2012articulated}
Pishchulin, L., Jain, A., Andriluka, M., Thorm{\"a}hlen, T., Schiele, B.:
  Articulated people detection and pose estimation: Reshaping the future.
  Computer Vision and Pattern Recognition (CVPR), 2012 IEEE Conference on pp.
  3178--3185 (2012)

\bibitem{qiu2016generating}
Qiu, W.: Generating human images and ground truth using computer graphics.
  Ph.D. thesis, University of California, Los Angeles (2016)

\bibitem{romero2015flowcap}
Romero, J., Loper, M., Black, M.J.: Flowcap: 2d human pose from optical flow.
  German Conference on Pattern Recognition pp. 412--423 (2015)

\bibitem{simonyan2014very}
Simonyan, K., Zisserman, A.: Very deep convolutional networks for large-scale
  image recognition. arXiv preprint arXiv:1409.1556  (2014)

\bibitem{stark2010back}
Stark, M., Goesele, M., Schiele, B.: Back to the future: Learning shape models
  from 3d cad data. Bmvc  \textbf{2}(4), ~5 (2010)

\bibitem{su2015render}
Su, H., Qi, C.R., Li, Y., Guibas, L.J.: Render for cnn: Viewpoint estimation in
  images using cnns trained with rendered 3d model views. Proceedings of the
  IEEE International Conference on Computer Vision pp. 2686--2694 (2015)

\bibitem{sun2016correlation}
Sun, B., Feng, J., Saenko, K.: Correlation alignment for unsupervised domain
  adaptation. arXiv preprint arXiv:1612.01939  (2016)

\bibitem{sun2015generating}
Sun, B., Peng, X., Saenko, K.: Generating large scale image datasets from 3d
  cad models. CVPR 2015 Workshop on The Future of Datasets in Vision  (2015)

\bibitem{sun2009multi}
Sun, M., Su, H., Savarese, S., Fei-Fei, L.: A multi-view probabilistic model
  for 3d object classes. Computer Vision and Pattern Recognition, 2009. CVPR
  2009. IEEE Conference on pp. 1247--1254 (2009)

\bibitem{varol2017learning}
Varol, G., Romero, J., Martin, X., Mahmood, N., Black, M.J., Laptev, I.,
  Schmid, C.: Learning from synthetic humans. 2017 IEEE Conference on Computer
  Vision and Pattern Recognition (CVPR 2017)  (2017)

\bibitem{wei2016convolutional}
Wei, S.E., Ramakrishna, V., Kanade, T., Sheikh, Y.: Convolutional pose
  machines. Proceedings of the IEEE Conference on Computer Vision and Pattern
  Recognition pp. 4724--4732 (2016)

\bibitem{yosinski2014transferable}
Yosinski, J., Clune, J., Bengio, Y., Lipson, H.: How transferable are features
  in deep neural networks? Advances in neural information processing systems
  pp. 3320--3328 (2014)

\bibitem{yu15lsun}
Yu, F., Zhang, Y., Song, S., Seff, A., Xiao, J.: Lsun: Construction of a
  large-scale image dataset using deep learning with humans in the loop. arXiv
  preprint:1506.03365  (2015)

\bibitem{zhou2017scene}
Zhou, B., Zhao, H., Puig, X., Fidler, S., Barriuso, A., Torralba, A.: Scene
  parsing through ade20k dataset. Proceedings of the IEEE Conference on
  Computer Vision and Pattern Recognition  (2017)

\end{thebibliography}
\end{document}